\documentclass{article}
\usepackage{ijcai23}

\usepackage{times}
\usepackage{soul}
\usepackage{url}
\usepackage[hidelinks]{hyperref}
\usepackage[utf8]{inputenc}
\usepackage[small]{caption}
\usepackage{graphicx}
\usepackage{amsmath}
\usepackage{amsthm}
\usepackage{booktabs}
\usepackage{algorithm}
\usepackage{algorithmic}
\usepackage[switch]{lineno}

\newtheorem{theorem}{Theorem}

\title{NeuralFastLAS: Fast Logic-Based Learning from Raw Data}

\author{
Theo Charalambous
\and
Yaniv Aspis\and
Alessandra Russo
\affiliations
Imperial College London
\emails
\{theo.charalambous18, yaniv.aspis17, a.russo\}@imperial.ac.uk
}

\usepackage{multirow}
\usepackage{tikz}
\usepackage{gincltex}
\usepackage{pgfplotstable}

\usepackage{textcomp}

\newcommand{\asp}[1]{\mbox{$\mathtt{#1}$}}
\DeclareMathOperator{\codeif}{\mathtt{:-} }
\DeclareMathOperator{\naf}{\;\mathtt{not}\;}

\newcommand{\NF}{NeuralFastLAS\ }
\newcommand{\MetaABD}{$Meta_{ABD}$}

\newcommand{\FLAS}{FastLAS\ }
\usepackage{theo_commands}

\begin{document}

\maketitle

\begin{abstract}
Symbolic rule learners generate interpretable solutions, however they require the input to be encoded symbolically. Neuro-symbolic approaches overcome this issue by mapping raw data to latent symbolic concepts using a neural network. 
Training the neural and symbolic components jointly
is difficult, due to slow and unstable learning, hence many existing systems rely on hand-engineered rules to train the network. 
We introduce NeuralFastLAS, a
scalable and fast end-to-end 
approach that trains a neural network jointly with a symbolic learner. For a given task, 
NeuralFastLAS computes a relevant set of rules, proved to contain an optimal symbolic solution, trains a neural network using these rules, and finally finds an optimal symbolic solution to the task while taking network predictions into account.
A key novelty of our approach is learning a posterior distribution on rules 
while training the neural network to improve stability during training.
We provide theoretical results for a sufficient condition on network training
to guarantee correctness of the final solution.
Experimental results demonstrate
that NeuralFastLAS is able to achieve
state-of-the-art accuracy in arithmetic and 
logical tasks, with a training time that is 
up to two orders of magnitude faster than other 
jointly trained
neuro-symbolic methods.

\end{abstract}

\section{Introduction}

Inductive Logic Programming (ILP) Systems \cite{metagol,ILASP} 
induce a set of first-order logical rules that 
can be used to explain observations. 
The solutions generated by ILP systems are naturally interpretable by humans, generalizable, and require very little data. However, ILP systems are limited to only learning from symbolic inputs. In many settings, data is instead presented in a raw form, such as a set of images or natural text. To overcome this issue, neuro-symbolic methods have been proposed in recent years. A promising approach involves separating the system into two different components: a neural component, which is responsible for mapping raw data to latent symbolic concepts, and a symbolic component which reasons over these latent concepts and produces a final answer. However, training the neural and symbolic components jointly is a difficult task. On the one hand, when the neural component is untrained, the symbolic component receives highly noisy input. On the other hand, the neural component does not have a reliable training signal before the symbolic component can induce the correct rules. For this reason, many neuro-symbolic systems use hand-engineered rules to train the neural network \cite{deepproblog,neurasp}.

To our knowledge, three approaches have been proposed that jointly train a neural network and use an ILP system to induce first-order rules. \MetaABD\ \cite{wangzhou_ijcai} builds on top of  Meta-Interpretive Learning \cite{MIL} and uses a process of abduction to derive the most likely rules and labels for use in backpropagation. However, it cannot learn rules involving default negation. NSIL uses ILASP \cite{ILASP} or FastLAS \cite{fastlas2} to learn answer set programs. At each iteration, it induces a hypothesis and trains the neural network using NeurASP \cite{neurasp}. It uses a process of exploration/exploitation of examples to improve learning. However, its dependence on the repeated search for an optimal hypothesis results in a slow training process and scalability issues. The Apperception Engine \cite{aperception} is another method that trains two components together, however the perception component is a binary neural network whose weights can be 0 or 1, hence it is unlikely to be able to scale to complex perception tasks.

In this paper, we introduce NeuralFastLAS, a neuro-symbolic learning system built on top of FastLAS. 
NeuralFastLAS receives examples in the form of raw data inputs together with a final label. 
From these examples, the system trains a neural network to recognise latent symbolic concepts 
from the raw data and induces an answer set program that can explain the final label given these 
latent concepts. Similarly to FastLAS, NeuralFastLAS first constructs an opt-sufficient subset 
of the hypothesis space that is guaranteed to contain an optimal hypothesis. Using this smaller 
set of rules, NeuralFastLAS trains a neural network to recognise the latent concepts using
a semantic loss function \cite{semantic_loss}, greatly speeding up training. To help guide the 
training, NeuralFastLAS uses a novel technique of learning a posterior distribution over the rules 
in the opt-sufficient subset. Once the neural network is trained, the final hypothesis is constructed 
by finding a set of rules of the shortest length that maximise the prediction scores of the network over the raw data.

We prove theoretical results for the correctness of the opt-sufficient subset produced by NeuralFastLAS and show a sufficient condition on the convergence of the neural network to guarantee that the correct rules are learnt. Furthermore, we evaluate NeuralFastLAS on arithmetic and logical tasks and show
that it consistently achieves state-of-the-art accuracy while
having training times that are orders of magnitude faster than
other neuro-symbolic systems that learn rules.

The paper is structured as follows. Section 2 provides background knowledge on answer set programming and the FastLAS system for learning such programs. Section 3 formalises the neuro-symbolic learning task that NeuralFastLAS solves. Section 4 covers each stage of the NeuralFastLAS algorithm in detail. Section 5 presents the results of a systematic evaluation of the NeuralFastLAS. Section 6 concludes the paper.

\section{Background}
In this section we introduce the concepts needed throughout the paper.~Let $\asp{h}$, $\asp{b_1,\ldots,b_n}$, $\asp{c_1,\ldots,c_m}$ be any atoms.~A \emph{normal rule} $r$ has the form $\asp{h \codeif b_1,\ldots, b_n,}$ ${\naf c_{1},\ldots, \naf c_{m}}$, where $\asp{h}$ is the \textit{head} of the rule (head(r)), $\asp{b_1,\ldots, b_n, \naf c_{1},\ldots, \naf c_{m}}$ is (collectively) the \textit{body} of the rule (body(r)), and ``$\naf$'' represents \textit{negation as failure}. 
A rule $r'$ is said to be a \textit{subrule} of $r$ (denoted $r' \leq r$) if $\head(r) = \head(r')$ and $\body(r') \subseteq \body(r)$. $r'$ is a \textit{strict subrule} if $r' \leq r$ and $r' \neq r$.
\textit{Constraints} are rules of the form $\asp{\codeif b_1, ..., b_n, c_1, ..., c_m}$, whereas \textit{choice rules} have the form $\asp{l\{h_1,\ldots,h_k\}u\codeif}$ $\asp{b_1,\ldots, b_n,\naf c_{1},\ldots, \naf c_{m}}$. The head $\asp{l\{h_1\ldots,h_k\}u}$ of a choice rule is called \textit{aggregate}, with $\asp{l}$ and $\asp{u}$ as (positive) integers and $\asp{h_1,\ldots,h_k}$ as any atoms. An ASP program is a set of normal rules, choice rules and constraints. The Herbrand Base of an ASP program $\Pi$, denoted $HB_{\Pi}$, is the set of all ground (with no variables) atoms formed using the predicates and constants that appear in $\Pi$. Subsets of $HB_{\Pi}$ are called \textit{interpretations} of $\Pi$. A ground (with no variables) aggregate $\asp{l\{h_1,\ldots,h_k\}u}$ is satisfied by an interpretation $I$, if  $\asp{l}\leq | I\cap\{\asp{h_1\ldots,h_k}\}|\leq \asp{u}$.
The semantics of an ASP program $\Pi$ is defined with respect to the \textit{reduct} of $\Pi$. Specifically, given an interpretation $I$ of $\Pi$, the \textit{reduct} of $\Pi$, denoted $\Pi^{I}$, is constructed from the grounding of $\Pi$ in four steps. The first two steps deal with the negation as failure ``$\naf$'' by i) removing all ground rules in $\Pi$ whose bodies contain the negation of an atom in $I$ and ii) removing all the negative literals from the remaining rules. The third step deals with the constraints and choice rules whose head is not satisfied by $I$ by replacing their respective (empty) head with $\bot$ (where $\bot\not\in HB_{\Pi}$). The final step deals with the remaining choice rules $\asp{\{h_1,\ldots h_m\}\codeif b_1,\ldots b_n}$ by replacing each of them with the set of rules
$\asp{\{h_i\codeif b_1,\ldots b_n\mid h_i\in I\cap
\{h_1,\ldots h_m\}\}}$. Any interpretation $I\subseteq HB_{\Pi}$ is an \textit{answer set} of $\Pi$ if it is the minimal model of the reduct $\Pi^{I}$. We denote with $AS(\Pi)$ the set of answer sets of $\Pi$. 

Since the \NF algorithm that we propose in this paper is built on top of the \FLAS  algorithm \cite{fastlas}, we briefly describe its main steps. \FLAS is a logic-based system for learning answer set programs capable of solving observational predicate learning (OPL) tasks over large search spaces. OPL tasks aim to learn concepts that are directly observable from given examples, and their solutions are ASP programs, called \textit{hypotheses}, that define such concepts. As is common in ILP systems, \FLAS uses mode declarations as language bias to define the search space (also called \textit{hypothesis space}) of a learning task. A mode bias $M$ specifies two sets, $M_h$ and $M_b$, of a mode head and body declarations, respectively. Informally, a mode declaration is an atom whose arguments are either $\asp{var(t)}$ or $\asp{const(t)}$, where $\asp{t}$ is a constant called a \textit{type}. An atom is compatible with a mode declaration $m$ if it uses the predicate of $m$ and replaces every argument $\asp{var(t)}$ in $m$ with a variable of type $\asp{t}$, and every $\asp{const(t)}$ in $m$ with a constant of type $\asp{t}$. Body declarations can be negated atoms using negation as failure.

\begin{definition}[Hypothesis space]
Let $M=\langle M_h, M_b\rangle$ be a mode bias. The hypothesis space $S_M=\{r_i | i\geq 1\}$ is the set of rules $r_i$ such that head($r_i$) is compatible with a mode head declaration, and each literal in body($r_i$) is compatible with a mode body declaration.
\end{definition}

\noindent
\FLAS learns ASP programs from a given set of examples, within the context of a (possibly empty) background knowledge. Examples are \textit{context-dependent partial interpretation} (CDPI). A CDPI $e=\langle e_{id}, e_{pi}, e_{ctx}\rangle$, where $e_{id}$ is the example identifier, $e_{pi}$ is a partial interpretation composed of a pair of disjoint sets of atoms $e_{pi}=\langle e^{inc},e^{exc}\rangle$ called the \textit{inclusions} and \textit{exclusions} respectively, and $e_{ctx}$ is an ASP program consisting of normal rules, called a \textit{context}. A program $\Pi$ \textit{accepts} a CDPI example $e$, if and only if there is an answer set $A$ of  $\Pi\cup e_{ctx}$ such that $e^{inc}\subseteq A$ and $A\cap e^{exc}=\emptyset$. We now recall the definition of a Learning from Answer Sets (LAS) task~\cite{fastlas}.

\begin{definition}[LAS task]
A Learning from Answer Set (LAS) task is a tuple $T=\langle B, M, E\rangle$ where $B$ is an ASP program, called \textit{background knowledge}, $E$ is a finite set of CDPIs, and $M$ is a mode bias. For any hypothesis $H\subseteq S_{M}$:
\begin{itemize}
\item For any $e\in E$, $H$ \textit{covers} $e$ iff $B\cup H$ accepts $e$.
\item $\mathcal{S}_{len}(H)$ is the number of literals in $H$, written $|H|$.
\item
$H$ is a solution of $T$ if $H$ covers all $e \in E$. $H$ is \textit{optimal}  if there is no solution $H^{'}$ such that $\mathcal{S}_{len}(H^{'})<\mathcal{S}_{len}(H)$.
\end{itemize}
\end{definition}

FastLAS is restricted to solving non-recursive, OPL tasks. A task is non-recursive if no predicate in $M_h$ occurs in $M_b$ or the body of any rule in $B \cup e_{ctx}$ for any $e \in E$.

To compute an optimal solution for a given LAS task $T$, \FLAS algorithm uses four main steps. Informally, in the first step it constructs a \textit{SAT-sufficient} space which is comprised of two sets $C^+(T)$ and $C^-(T)$; in the second step, called \textit{generalisation}, it constructs $G(T)$ as a set of rules that generalise $C^+(T)$ without covering any rules in $C^-(T)$. It then \textit{optimises} $G(T)$, in its third step, into an \textit{OPT-sufficient} hypothesis space. The final \textit{solving} step, uses the OPT-sufficient hypothesis space to compute an optimal solution.

\section{Neural-Symbolic Learning}

Many neural-symbolic architectures 
\cite{wangzhou_ijcai,neurasp} consist of distinct 
perception and reasoning components. The perception component maps 
the raw data to (latent) symbolic concepts while the reasoning component 
finds a final label using symbolic background knowledge. To train both components, a dataset $\mathcal{D}$ is available, composed of pairs $(x,y) \in \X \times \Y$ where $\X$ is the set of all possible raw inputs, $\Y$ the set of final labels, and the latent symbolic concepts are unlabelled. We denote the space of latent symbolic concepts as $\mathcal{Z}$.


Formally, the perception model is parameterised by $\theta$. For a given raw 
data input, the perception model 
outputs a probability $P_\theta (z | x) = P(z | \theta , x)$ for each $z \in \Z$. The symbolic component consists of background knowledge $B$ and a hypothesis $H$, which are used together to map the symbolic concepts in $z$ to the labels $y$. To optimise both $\theta$ and $H$ jointly, we use the following objective function \cite{wangzhou_ijcai}:
\begin{equation}
    \label{eq:max1}
    (H^*, \theta^*) = \argmax_{H, \theta} \prod_{(x, y) \in \D} \sum_{z \in \Z} P(y, z | B, x, H, \theta)
\end{equation}
Note that this can be rewritten as
\begin{equation}
    \label{eq:sub1}
    P(y, z | B, x, H, \theta) = 
    P(y | B, H, z) P_{\sigma^*} (H | B) P_\theta (z | x)
\end{equation}
where $P_{\sigma^*}$ is the Bayesian prior distribution on 
first-order logic hypothesis \cite{prior}. To solve for the objective function above in the context of FastLAS, we propose the NeuralFastLAS algorithm.

\subsection{NeuralFastLAS Learning Tasks}

This section formulates the learning tasks used within the NeuralFastLAS
architecture. It differs from the learning tasks in FastLAS
as the context of an example is not symbolically given, but consists of raw data.

\begin{definition}
    A raw-data example is of the form $e = \langle \eid, \langle \epiinc, \epiexc \rangle, \eraw \rangle$. $\eid$ is the identifier for the example,
    $\epiinc$ and $\epiexc$ are the inclusion and exclusion set, respectively, and
    $\eraw$ is the set of raw data for that example. 
\end{definition}

An example of raw data may be a set of images, the corresponding latent symbolic 
labels may be object names or numbers that appear in the image.
For any data point $d = (x, y) \in \D$, we generate the example 
$e_d = \langle d_{id}, \langle \{ y \}, \Y \setminus \{ y \} \rangle, x \rangle$ 
where $d_{id}$ is any unique identifier.

\begin{definition}
    A NeuralFastLAS task is of the form $T = \langle B, M, E, \Z, \Theta \rangle$ 
    where $B$ is the background knowledge in the form of an ASP program,
    $M$ is the mode bias, $\Z$ is the latent space and $E$ is a set 
    of raw-data examples. $\Theta$ is the space of neural network parameters 
    such that $\theta \in \Theta$ maps $x \in \X$ to $\Z$.
\end{definition}

A solution for a NeuralFastLAS task is a pair $(H, \theta)$ where $H \subset S_M$ and $\theta \in \Theta$ is the neural network's parameters. 
An optimal solution of the task is a solution of equation \ref{eq:max1}.
As NeuralFastLAS is built on top of the FastLAS algorithm, it inherits the restriction to non-recursive, OPL tasks.

\section{NeuralFastLAS Algorithm}

This section presents the NeuralFastLAS algorithm. It is 
built on top of the FastLAS system \cite{fastlas2}. There are six steps 
to the NeuralFastLAS system: (1) Abduction; (2) Construction of the SAT-sufficient
subset; (3) Generalisation; (4) Optimisation; (5) Neural training; (6) Solving.

Steps (1) to (4) construct the opt-sufficient subset of the task,
that is the set of rules that cover at least one example for 
some valid combination of latent labels, and
does not violate any example. The opt-sufficient subset is 
then used to train the neural network using the semantic loss
function \cite{semantic_loss}
in step (5). Finally, using predictions from the neural network,
we find the optimal solution with respect to \autoref{eq:sub1}
in step (6).

\subsection{Abduction}

The abduction stage computes the set of valid network outputs for each  $e \in E$ while taking into account any constraints given by $B$. 
To do so, a symbolic program $P_{\Z}$ is constructed whose answer sets correspond to all possible neural network choices. $\Z$ is of the 
form $\Z = \Z_1 \times \Z_2 \times ... \times \Z_n$
where the pipeline takes $n$ raw data inputs per data point.
For each $\Z_i$, we add the choice rule 
\begin{equation}
    \asp{1\{nn(i, \Z_{i,1}), ..., nn(i, \Z_{i,m}) \}1.}
\end{equation}
to $P_{\Z}$. The answer sets of $P_{\Z}$ correspond to every combination 
of latent labels.
The predicate \texttt{nn/2} is a reserved name in
NeuralFastLAS. It
maps a raw data identifier to the corresponding symbolic label of the raw data.
For example, if the first raw data input is an image of a cat, then 
the final ASP program will contain the atom \texttt{nn(1, cat)}.

The set of valid neural network outputs for each example 
corresponds to the answer
sets $AS(B \cup P_{\Z})$. We refer to each such answer set as a \textit{possibility} and denote $\text{poss}(e) = AS(B \cup P_{\Z})$.
With this, we can define coverage in the context 
of NeuralFastLAS:

\begin{definition}
    We say that a hypothesis $H$ covers a NeuralFastLAS 
    example $e$ if there exists a possibility $A \in AS(B \cup P_{\Z})$
    such that $A \cup H$ accepts $e$. 
    
    The set of neural networks 
    outputs that can be used by $H$ to cover an example $e$ 
    is denoted $\cov(H, e)$. Formally, $\cov(H, e) = \{ A \cap \Z : A \in AS(B \cup P_{\Z}), A \cup H \text{ accepts } e \}$.
\end{definition}

\subsection{Computing the SAT-Sufficient Subset and Generalisation}

After computing the symbolic possibilities, NeuralFastLAS
produces a SAT-sufficient space by first constructing the sets  
$C^+(T, p)$ and $C^-(T, p)$ of maximal
rules that prove at least one literal in the inclusion or
exclusion sets of a possibility $p$, respectively.
$C^+(T) = \cup_{e \in E} \cup_{p \in \text{poss}(e)} C^+(T, p)$ 
is the set of all
SAT-sufficient rules. $C^-(T)$ is defined analogously.


The abduction stage of NeuralFastLAS can produce a large 
number of possibilities. 
With many examples, the number of possibilities 
can make the task computationally intensive.
To explore the search space intelligently, 
NeuralFastLAS
prunes $S_M$ by taking into account
constraints over the mode bias.
Two key constraints of NeuralFastLAS are:
\begin{enumerate}
    \item \textbf{Symmetry of 
Predicate Arguments}: If the arguments \texttt{A} and \texttt{B}
are marked as symmetric in the predicate \texttt{h(A, B, C)}, then
rules that differ only in the order of \texttt{A} and \texttt{B} 
are considered the same and only appear once in the SAT-sufficient
subset.

    \item \textbf{Capture and Release Arguments:} Every argument in every literal is referred to as 
    either a ``capture" argument (denoted with $^-$) or a ``release" argument (denoted with $^+$). 
    Intuitively, a 
    rule conforms to capture and release conditions if: (1) For any
    body literal with a capture argument, the variable in 
    that argument must also be present in a release 
    argument of another body literal or the head literal.
    (2) For variables 
    in the capture arguments 
    in the head literal, that variable must also be present as 
    a release argument of a body literal.
    (3) Every variable present
    in the release argument of a literal must be unique. Consider $M_h = \{ f(var(t^+), var(t^-) \}$ and 
    $M_b = \{ g(var(t^-), var(t^+)), h(var(t^-), var(t^+)) \}$,
    then the rule $f(A, B) \codeif g(A, C), h(C, B)$ conforms to 
    the capture and release variables, whereas the rule 
    $f(A, B) \codeif g(A, C), h(D, C)$ does not as the release arguments of 
    $g$ and $h$ are not unique.
\end{enumerate}

A formal description of symmetry of 
predicate arguments and capture and release variables is 
given in the supplementary material.

The generalisation step is identical to that in the original FastLAS
system. Each rule from the SAT-sufficient subset 
is generalised and added to the new set 
$G(T)$. 

\subsection{Optimisation}

At this stage, we have the set of generalised rules $G(T)$. 
We proceed by defining the 
following notion of \textit{neural optimality}, which aims to 
find the most relevant subrules of $G(T)$ that do not violate
all of the possibilities of any example.
Note that in NeuralFastLAS, every example must be covered by making 
an appropriate choice for the latent labels.

\begin{definition}[Neural Optimisations]
  \label{def:neuralopt}
  Given a rule $r$, we define the \textit{optimisations of 
  $r$}, denoted $\neuropt(r)$, to be the set of 
  rules $r'$ that satisfy the following 
  conditions:
  \begin{enumerate}
    \item $r'$ is a subrule of $r$.
    \item There does not exist a rule $r''$ such that 
    $r'' < r'$ and $r'$ and $r''$ cover the exact same 
    possibilities.
    \item \label{item:neuroopt1} There does not exist any example $e \in E$
    such that for every $p \in \text{poss}(e)$,
    $r'$ is a subrule of a rule in $C^-(T, p)$.
    \item $r'$ is a smallest such rule  
    satisfying (1), (2) and (3).
  \end{enumerate}
  The opt-sufficient subset is then defined as $S_M^\text{opt}$
  where 
  $$
S_M^\text{opt} = \bigcup_{r \in G(T)} \neuropt(r)
  $$
\end{definition}

\Cref{item:neuroopt1} is a
pruning criterion introduced in NeuralFastLAS.
Intuitively, \cref{item:neuroopt1} says that 
if a rule violates every possibility of an example,
then it is guaranteed to not be included in the final 
solution so we can prune it at the optimisation 
stage.

\subsection{Neural Training}

The optimisation stage produces the opt-sufficient subset of 
the task $T$. With this, we can train the neural network.
The neural network training stage of NeuralFastLAS consists of:
\begin{enumerate}
  \item \textbf{Training the Perception Component:}
  The raw data is fed into a neural network 
  which outputs probability vectors predicting 
  the corresponding latent label for each 
  raw data input. During the gradient descent step,
  NeuralFastLAS optimises the parameters to improve 
  the accuracy of the network.
  \item \textbf{Learning a Posterior Distribution on the Rules:}
  NeuralFastLAS also uses an extra parameter
  $\theta_R$ to represent a score for the rules in $\smopt$. 
  This is used to improve stability when training the network.
\end{enumerate}
The neural network 
has two heads: the first is the perception
component, and the second is the rule probability 
predictor. The output of the neural network 
is $\theta_R$ concatenated with
all of the $P_\theta(z_i | x_{i})$ for $x_i \in \eraw$.

\begin{figure}[!tb]
    \centering
        \begin{tikzpicture}[scale=0.425, trim left=-2cm]
      \begin{scope}[shift={(-0.5, 0)}]
      \node[align=center, text width=3.21cm] at (-2.45,2.5) 
        {\fontsize{8}{8}\selectfont Input\\$x \in \X$};
      \draw [fill=white] (-3,4) rectangle (-1.5, 5.5);
      \draw [fill=white] (-3.2,3.8) rectangle (-1.7, 5.3);
      \draw [fill=white] (-3.4,3.6) rectangle (-1.9, 5.1);
    \end{scope}

    \draw[->, thick] (-1.5,4.55) -- (-0.5,4.55) {};

      \begin{scope}[shift={(-0.75, 0.2)}, scale=0.9]
        \draw[fill=lightgray] (4,3.5)--(1,3.5)--(1,5.6)--(4,5.6)--(4,3.5);
        \draw[fill=lightgray] (1,5.6)--(4,5.6)--(4.5,6.1) -- (1.5,6.1) -- (1,5.6);
        \draw[fill=lightgray] (4.5,6.1) -- (4.5,4) -- (4,3.5) -- (4,5.6) -- (4.5,6.1);
      \end{scope}
      \node[align=center, text width=1.8cm] at (1.5,4) 
        {\fontsize{8}{8}\selectfont Neural Network\\ \null};

      \draw[->, thick] (3.8,4.7) -- (4.8,4.7) {};
      \draw[->, thick, dashed] (4.8,4.4) -- (3.8,4.4) {};

      \node[align=center, text width=3.21cm] at (5.75,2.5) 
        {\fontsize{8}{8}\selectfont Latent Predictions\\$P_\theta (z | x)$};
      
      \begin{scope}[shift={(-1, 0)}]
        \draw [fill=white] (6.7,4.0) rectangle (7.0, 5.5);
        \draw [fill=white] (6.5,3.8) rectangle (6.8, 5.3);
        \draw [fill=white] (6.3,3.6) rectangle (6.6, 5.1);
      \end{scope}

      \draw[->, thick] (6.5,4.7) -- (7.8,5.33) {};
      \draw[->, thick, dashed] (7.8,5.03) -- (6.5,4.4) {};

      \draw [decorate,decoration={brace,amplitude=5pt},yshift=0pt]
      (8.3,3.85) -- (8.3,6.5) node [black,midway,xshift=-0.6cm] 
      {};
      \draw [decorate,decoration={brace,amplitude=5pt},yshift=0pt]
      (8.3,6.55) -- (8.3,7.85) node [black,midway,xshift=-0.6cm] 
      {};

      \draw [fill=white] (8.5,3.85) rectangle (8.8, 7.85);

      \draw [fill=white] (8.5,6.5) -- (8.8, 6.5);
      \draw [fill=white] (8.5,4.51) -- (8.8, 4.51);
      \draw [fill=white] (8.5,5.84) -- (8.8, 5.84);
      \node[text centered] at (8.65,5.3) {$\vdots$};

      \draw[->, thick] (6,7.35) -- (7.8,7.35) {};
      \draw[->, thick, dashed] (7.8,7.05) -- (6,7.05) {};

      \node[align=center, text width=3.21cm] at (8.65,8.6) 
        {\fontsize{10}{10}\selectfont $p$};
      
      \begin{scope}[shift={(-0.5, 0)}]
        \draw [fill=white] (5.5,6.45) rectangle (5.8, 7.95);
        \node[align=center, text width=3.21cm] at (5.65,8.5) 
          {\fontsize{8}{8}\selectfont $P(\smopt | \theta_R)$};

        \draw[->, thick] (3.3,7.35) -- (4.8,7.35) {};
        \draw[->, thick, dashed] (4.8,7.05) -- (3.3,7.05) {};
        
        \begin{scope}[shift={(-0.5, 0)}]
          \draw [fill=lightgray] (2.8,6.45) rectangle (3.1, 7.95);
          \node[align=center, text width=3.21cm] at (2.95,8.5) 
            {\fontsize{10}{10}\selectfont $\theta_R$};
        \end{scope}
      \end{scope}

      \draw[->, thick] (9.2,6) -- (10.2,5) {};
      \draw[->, thick, dashed] (10.2,4.7) -- (9.2,5.7) {};

      \begin{scope}[shift={(0, -1.0)}]
        \node[align=center, text width=2cm] at (12.25,5.85) 
              {\fontsize{8}{8}\selectfont Semantic Loss Function};

        \draw[->, thick] (12,4.8) -- (12,3.5) {};
        \draw[->, thick, dashed] (12.5,3.5) -- (12.5,4.8) {};

        \node[align=center, text width=2cm] at (12.25,2.85) 
              {\fontsize{8}{8}\selectfont $L^s(y, p)$};
        
        \draw[->, thick] (12.25,7.75) -- (12.25,6.75) {};
        \node[align=center, text width=2cm] at (13.25,7.25) 
              {\fontsize{8}{8}\selectfont $\AS_y$};
        \node[align=center, text width=2cm] at (12.25,8.25) 
              {\fontsize{8}{8}\selectfont ASP Solver};

        \draw[->, thick] (12.25,9.75) -- (12.25,8.7) {};
        \node[align=center, text width=2cm] at (12.25,10.25) 
              {\fontsize{8}{8}\selectfont Label $y \in \Y$};
      \end{scope}

      \draw (-1,1) rectangle (9.5, 10);
      \draw (15,3.25) rectangle (10, 8.25);
    \end{tikzpicture}
    \caption{Schematic of the Neural Component of 
      NeuralFastLAS. Gray sections show where gradient descent 
      steps occur. 
      The dashed line show how gradients are propagated. 
      The two boxes denote the neural component (left) and 
      semantic component (right).}
    \label{fig:fl_neural}
\end{figure}
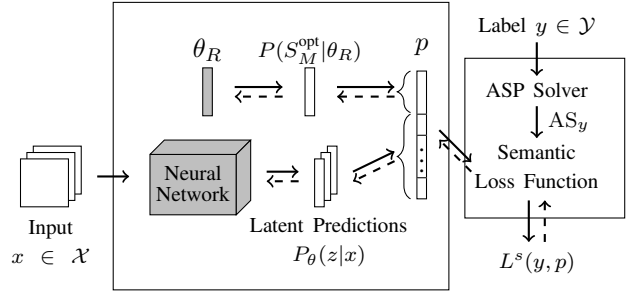

The semantic loss function
is intended to capture how close a neural 
network's predictions are 
to satisfying some symbolic constraints.
NeuralFastLAS uses the
semantic loss function to find the latent 
label predictions and rules in $\smopt$
can be used to prove the downstream label 
and propagate this information back 
to the neural network during gradient descent.
To compute the semantic loss for an example $e$, the answer set program $P_e = B \cup P_{\Z} \cup \smopt \cup \{ \asp{\codeif \naf y.} \} \cup \asp{1 \{ use(0..N) \} 1.}$ is constructed.  
Each rule $r_i \in \smopt$ is augmented with a literal \asp{use(i)} to indicate if it is used in an answer set when proving the downstream label.
This way, each answer set of $P_e$ corresponds to a choice of the neural network labels and the rule that entails $y$.



\begin{definition}[Semantic Loss in ASP]
  \label{def:asp_semantic}
  Let $P_e$ be the program for which we are 
computing the semantic loss. 
For a final label $y$, let $\AS_y$ denote the set of answer sets of 
$P_e$ that contain $y$ and
$X$ be the set of all possible ground \asp{nn} and \asp{use} facts.
We define the semantic loss as 
\begin{equation}
  L^s (y, p) = - \log \sum_{A \in \AS_y}\  \prod_{\substack{i \in \{ 0, ..., n\}\\X_i \in A \cap X}} p_i  \prod_{\substack{i \in \{ 0, ..., n\}\\X_i \in A^c \cap X}} (1 - p_i)
\end{equation}
where $p$ is a posterior probability vector given as output by the neural network and
$p_i$ denotes the predicted probability
of atom $X_i$. 
\end{definition}

Previous symbolic reasoners \cite{efy} used 
the semantic 
loss function with handwritten rules. NeuralFastLAS is different in that it learns rules and the network is trained before the final hypothesis is found. This introduces a new 
problem: during training, the many incorrect rules produce a lot 
of noise in the loss computation.
To compensate for this noisy signal, we  learn a probability distribution on the 
rules in $\smopt$.
We do this by introducing the parameter
$\theta_R \in \mathbb{R}^{\left|\smopt\right|}$ such that
  $P(r_i | \theta_R) = \softmax(\theta_R)$
  gives a probability vector.
Intuitively, 
  we can see $P(r_i | \theta_R)$ as 
  a learnt probability score that
  rule $r_i$ from $\smopt$ will cover a data point.
  


\subsection{Solving}

After the network has been trained, 
the solving stage constructs the program
$P_\text{solve}$ to compute an optimal hypothesis thus solving the symbolic 
task. In NeuralFastLAS, the following prior is introduced
\begin{equation}
    P_{\sigma^*} (H | B) = (e-1) e^{- \abs{H}}
\end{equation}
in order to perform a $\log$ transformation
\begin{align}
    &\argmax_{H, \theta} P(y | B, H, z) P_{\sigma^*} (H | B) P_\theta (z | x)
    \\ &= \argmin_{H, \theta} \left[ \abs{E} \abs{H} 
    + \sum_{e \in E} \min_{z \in \cov(H, e)} - \log P_\theta (z | \eraw ) \right]
    \label{eq:optimal}
\end{align}

A solution $(H^*, \theta^*)$ is optimal if it satisfies \autoref{eq:optimal}.
There are two distinct parts of the equation:
the first is a value that scores the hypothesis based on its length,
the second is a value that scores the best 
network choice $z$ 
out of all possible network choices that can be used to cover $e$ with $H$.


The solving stage involves constructing a program $P_\text{solve}$ that includes an ASP representation of equation~\ref{eq:optimal} in terms of weak constraints (details of the construction of $P_\text{solve}$ can 
be found in the supplementary material).  
An optimal answer set of $P_\text{solve}$ corresponds to an optimal hypothesis with respect to a given $\theta$. 
We denote this solution by 
$H = \neural_solve(T, \theta)$.

\subsection{Optimality of NeuralFastLAS}
 
We prove two key theoretical properties of the NeuralFastLAS algorithm. 
Let $\theta^*$ represent the perfect neural network, then we define the
notion of a \textit{correct hypothesis} $H^*$ to be an optimal hypothesis 
with respect to $\theta^*$, $H^* = \neural_solve(T, \theta^*)$.
First, we show that the opt-sufficient subset contains a correct hypothesis for a task $T$.
Second, we prove a sufficient condition on the convergence of the neural network 
to guarantee that the solving stage returns a correct hypothesis.

\begin{theorem}
    Let $T$ be a NeuralFastLAS task, then there exists a hypothesis 
    $H \subseteq \smopt$ in the opt-sufficient subset such that 
    $(H, \theta^*)$ is a solution to $T$ and for any other hypothesis $H'$
    such that $(H', \theta^*)$ is also a solution of $T$, $\mathcal{S}_{len}(H) \leq \mathcal{S}_{len}(H')$.
\end{theorem}

Since the solving stage follows the training of the neural network,
the results from the solving stage are dependent on the convergence of the network. 
We prove that if the network 
trains ``almost perfectly", then NeuralFastLAS 
returns a correct solution. It is important 
to emphasise that the condition of the network 
training ``almost perfectly" is \textbf{a sufficient, 
but not a necessary condition} for returning a correct hypothesis. 
The neural network can achieve suboptimal results but 
NeuralFastLAS may still find the optimal 
hypothesis.

\begin{definition}[Almost Perfectly Trained Networks]
  \label{def:almost_perf}
  A network with parameters $\theta$
  is said to have trained 
  \textit{almost perfectly} for a NeuralFastLAS task 
  $T$
  if the following bound is satisfied:
  Let $H^*$ be a correct hypothesis for 
  $T$, then for any input $x$ 
  with ground truth latent label $z^\text{gt}$,
  we have 
\begin{equation}
  \label{eq:perfectnn}
  P_\theta (z^\text{gt} | x)
  \geq 
  \frac{e^{\abs{E} \abs{H^*}}}{1 + e^{\abs{E} \abs{H^*}}}
\end{equation}
\end{definition}

The following theorem states that in the case of an almost perfectly
trained network, NeuralFastLAS always returns a 
correct solution.

\begin{theorem}
  \label{thm:nfl_correctness}
  Let $\theta$ be the parameters of an almost perfectly 
  trained network, then $H = \neural_solve(T, \theta)$
  is a correct solution of $T$.
\end{theorem}

\section{Experiments}

In this section, we discuss the empirical evaluation of NeuralFastLAS. We evaluate NeuralFastLAS on tasks involving arithmetic computations and logical reasoning\footnote{Code to be made available if the paper is accepted.}. 
The experiments conducted aim to answer the following questions:
(1) Can NeuralFastLAS learn a correct hypothesis and train the neural network jointly with high accuracy? 
(2) How does NeuralFastLAS 
compare to fully neural and other neuro-symbolic learning methods in terms of accuracy
and training time?
(3) Does the posterior 
distribution learnt over the rules accurately reflect 
the correct hypothesis? 
(4) Is NeuralFastLAS able to scale up with a larger hypothesis space? 

\begin{table*}[!h]
    \centering
    \begin{tabular}{l@{\extracolsep{4pt}}ccccc@{}}
        \toprule
        &&\multicolumn{4}{c}{End-to-End Accuracy} \\
        \cline{3-6} \\
        &Task Type&$a+b+c$&$a \times b+c$&$a \times b \times c$&E9P \\
        \midrule
CNN&&40.76 $\pm$ 1.71 \%&52.16 $\pm$ 1.51 \%&68.68 $\pm$ 1.14 \%&95.13 $\pm$ 0.43 \%\\
CBM&&90.66 $\pm$ 1.03 \%&93.52 $\pm$ 0.62 \%&95.65 $\pm$ 0.43 \%&97.69 $\pm$ 0.38 \%\\
NALU&&0.03 $\pm$ 0.02 \%&0.00 $\pm$ 0.00\%&0.19 $\pm$ 0.09\%&- \\
        \midrule
MetaABD&\multirow{3}{*}{Small}&33.03 $\pm$ 25.98 \%&93.82 $\pm$ 0.62 \%&94.45 $\pm$ 0.44 \%&-\\
NSIL&&8.02 $\pm$ 0.47 \%&86.44 $\pm$ 27.25 \%&96.22 $\pm$ 0.43 \%&79.93 $\pm$ 33.88 \%\\
NeuralFastLAS&&\textbf{96.33 $\pm$ 0.75 \%}&\textbf{96.48 $\pm$ 0.23 \%}&\textbf{96.81 $\pm$ 0.23 \%}&\textbf{97.72 $\pm$ 0.26 \%}\\
        \midrule
MetaABD&\multirow{3}{*}{Large}&\multicolumn{3}{l}{------------------------- Training Timeout (24 hours) ---------}&-\\
NSIL&&\multicolumn{4}{r}{------------------------- Training Timeout (24 hours) ----------------------------------}\\
NeuralFastLAS&&96.14 $\pm$ 0.45 \%&96.47 $\pm$ 0.29 \%&96.62 $\pm$ 0.21 \%&97.39 $\pm$ 0.61 \%\\
        \bottomrule
\end{tabular}
    \caption{Benchmark results for the four tasks. For each result, we show the mean and standard deviation of the final accuracy on the test set. A dash indicates that the experiment was not run for that model.}
    \label{table:accuracyresults}
\end{table*}

\paragraph{Experiments:} We answer these questions with 
four experiments: The first three tasks involve learning 
arithmetic formulae, specifically the formulae $a + b + c$,
$a \times b \times c$ and $a \times b + c$, where $a$, $b$ and $c$ are
given as MNIST images representing single-digit integers.
The final labels for each task appear in the background knowledge and are in the ranges 0-27, 0-90 and 0-729, respectively.
These tasks are challenging due to the large 
number of neural possibilities and the large hypothesis space;
previous neuro-symbolic methods that learn rules have only performed arithmetic 
experiments on two digits or with only addition and
multiplication in the search space.
\begin{table}
    \centering
    \begin{tabular}{lcc}
        \toprule
        &Small&Large\\
        \midrule
        Arithmetic&$+$, $\times$&$+$, $\times$, $-$, \textsuperscript{$\wedge$}, $//$\\
        E9P&\texttt{even}, $+9$&\texttt{even}, $+1$, $+2$, ..., $+9$\\
        \bottomrule
    \end{tabular}
    \caption{
    Search space for each category of task. $//$ represents integer division.}
    \label{table:searchspace}
\end{table}

The last task is Even9Plus (E9P) \cite{nsil}. In this task, the input 
is two images representing single-digit integers. If the 
first digit is even, the result is the value of the second 
digit plus nine. Otherwise, the result is the value of the 
second digit. This task requires learning a rule with negation as failure.

To investigate how each method scales with the size of the search space,
we create two versions of each task. \autoref{table:searchspace} 
shows the body predicates that form the search space for each category of task.

\paragraph{Baselines:} We compare 
NeuralFastLAS against a variety of other methods. First, we 
compare NeuralFastLAS to fully neural methods: a CNN baseline, Concept Bottleneck Models (CBM) \cite{cbm} and 
Neural Arithmetic Logic Units (NALU) \cite{nalu}. The NALU is excluded from the E9P task as it is not 
designed for logical reasoning.
We also compare to two other neuro-symbolic learning models 
that jointly train the neural and symbolic components:
\MetaABD\footnote{We use the \MetaABD\ code available at \url{https://github.com/AbductiveLearning/Meta_Abd}} \cite{wangzhou_ijcai} and NSIL\footnote{The code is not publicly available, but we were given access after contacting the authors of \cite{nsil}.} \cite{nsil}.
The same CNN network was used for the neural component of each of the neuro-symbolic methods.
Note that \MetaABD\ cannot learn negation as failure and hence is unable to solve the E9P task. 

\paragraph{Experiment Setup:} Each experiment was run 10 times. Hyper-parameter tuning was carried out for each model
using a held-out validation set.
For the neuro-symbolic
models, we test against both forms of the mode bias.
All experiments were performed on a machine with the following specification: Ubuntu 22.04.1 LTS with an Intel\textsuperscript{\textregistered} Xeon\textsuperscript{\textregistered} W-2145 CPU @ 3.70GHz, an NVIDIA GeForce RTX 2080 Ti GPU and 64GB of RAM.

\subsection{Learning Arithmetic Formulae}

\begin{table*}[!tb]
    \centering
    \begin{tabular}{lccccc}
        \toprule
        &&\multicolumn{4}{c}{Training Time (minutes)} \\
        \cline{3-6} 
        &Task Type&$a+b+c$&$a \times b+c$&$a \times b \times c$&E9P \\
        \midrule
\MetaABD&\multirow{3}{*}{Small}&245.4&31.7&52.5&-\\
NSIL&&18.2&46.8&25.0&13.7\\
NeuralFastLAS&&\textbf{0.8}&\textbf{1.7}&\textbf{7.2}&\textbf{0.3}\\
        \midrule
\MetaABD&\multirow{3}{*}{Large}&\multicolumn{3}{r}{-- Training Timeout ( $>$ 24 hours) --}&-\\
NSIL&&\multicolumn{4}{c}{-- Training Timeout ( $>$ 24 hours) ----------}\\
NeuralFastLAS&&5.9&10.5&27.6&3.3\\
        \bottomrule
    \end{tabular}
    \caption{Mean training time of the neuro-symbolic systems}
    \label{table:traintimetable}
\end{table*}
The results of this experiment are summarised in \autoref{table:accuracyresults}. First, we discuss the answer to questions (1) and (2). NeuralFastLAS was able to correctly solve all the given tasks and achieve high accuracy. 
It was able to do so even though the neural network was not trained ``almost perfectly", showing that the condition in \autoref{def:almost_perf} is not a necessary one.

The fully neural models consistently perform worse than their 
neuro-symbolic counterparts. This is likely because 
the fully neural models are unable to utilise
any background knowledge. In particular, NALU struggles 
to learn. RNNs generally require many more training samples
due to problems propagating gradients \cite{rnn_difficult}.

\MetaABD\ is less consistent than NeuralFastLAS in its learning, as seen in the 
large standard deviation of accuracy. In the $a+b+c$ task,
the minimum accuracy \MetaABD\ achieves is 2.6\% while the highest is 82.0\%. 
On the other hand, NeuralFastLAS consistently 
achieves over 95.0\%. This demonstrates that NeuralFastLAS is
less sensitive to its initial conditions.
\MetaABD\ utilises a process of abduction to train the neural 
network, choosing the latent label with the highest probability. Hence, during the backpropagation stage of each example,
only the information of one label is used to train the network.
In the case that the wrong label is chosen, \MetaABD\ can 
get stuck in a local minimum, which explains the sensitivity to
initial conditions. On the other hand, NeuralFastLAS' use 
of the semantic loss allows it to take into account every
possible latent labelling during backpropagation, avoiding this problem.

The \MetaABD\ tasks are formulated in such a way that the equations it 
can learn are of the form $(a \circ_1 b) \circ_2 c$ where $\circ$ represents
an operator, so the search space is small. However, \MetaABD\ cannot solve the large arithmetic task within a reasonable
amount of time. This is because it has to recompute the 
abduced labels every epoch, whereas NeuralFastLAS computes the 
opt-sufficient subset once during the whole training pipeline.

NSIL fails to solve the $a+b+c$ task. This is because it must commit 
to a hypothesis before training the neural network. The range 
of $a+b+c$ is a subset of the ranges of many other formulae (e.g. $a \times a + b$) and so if the wrong
hypothesis is chosen in the first iteration, the network will train
incorrectly and become stuck in a local minimum.

\begin{table}[!tb]
    \centering
    \begin{tabular}{lccc}
        \toprule
        &\multicolumn{3}{c}{Latent Label Accuracy} \\
        \cline{2-4}
        &$a+b+c$&$a \times b + c$&$a\times b \times c$\\
        \midrule
        \MetaABD&54.0\%&97.3\%&97.4\%\\
        NSIL&12.2\%&96.8\%&88.6\%\\
        NeuralFastLAS&\textbf{98.8\%}&\textbf{98.7\%}&\textbf{98.5\%} \\
        \bottomrule
    \end{tabular}
    \caption{
    The latent label accuracy for the neuro-symbolic systems
    }
    \label{table:latentacc}
\end{table}

\autoref{table:latentacc} shows the latent accuracy achieved by both methods 
- despite the training of the network being weakly supervised 
through the use of semantic loss, NeuralFastLAS achieves an accuracy comparable 
to that of a supervised network training on MNIST labels.

\subsection{Learning with Negation-as-Failure}

In the E9P task, which requires learning rules with negation-as-failure, NeuralFastLAS 
outperforms NSIL. NSIL can become trapped in a local minimum if it chooses the wrong
hypothesis in the first iteration, but in most cases, both method achieve
similar a final accuracy. This similarity 
is due to both utilising the semantic loss to train the network. NeuralFastLAS uses it explicitly in 
its training stage, while NSIL uses it implicitly through its use of 
NeurASP \cite{neurasp}. 

However, NSIL fails to train within the time limit in the larger task.
The exploration/exploitation mechanism produces many noisy examples
which creates an incredibly complex optimisation task that is passed
to FastLAS. The large number of examples in combination with the large 
search space create a task that FastLAS struggles to solve quickly.

\subsection{The Posterior Rule Distribution}

\begin{figure}[!tb]
    \centering
    \includegraphics{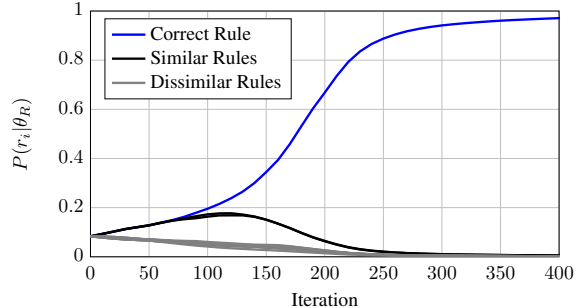}
    \caption{The learnt posterior distribution of the rules during the first quarter of the first training epoch for the task learning $a \times b + c$. The blue line represents the correct rule $a \times b + c$, the black lines represent 
    rules that are ``close" (namely $a \times c + b$ and $b \times c + b$.)}
    \label{fig:rules}
\end{figure}

To answer question (3), we present an example of how the 
posterior distribution of a NeuralFastLAS changes during 
the training of the network for the $a \times b + c$ task.
From \autoref{fig:rules}, we see that the algorithm successfully
learns the correct rule. The intuitive reason is that a ground truth
rule (that is, a rule used to generate the dataset) will be present
in at least one answer set when computing the answer sets for the 
semantic loss. In turn, the backpropagation will favour this rule 
over rules that appear less frequently in answer sets.

By the end of this epoch, the answer sets that use the correct rule from
the opt-sufficient subset will contribute to most of the semantic loss' value.
Hence the noise from the other answer sets is reduced and learning is 
much more stable. NeuralFastLAS  is 
able to fine-tune on the correct
rule before even reaching the solving stage, learning to better predicted probabilities to
use in solving.

\subsection{Training Time}
\label{subsection:trainingtime}

We turn our attention to question (4). \autoref{table:traintimetable} shows the training time 
for each of the models.
We look at the $a+b+c$ task as an example, NeuralFastLAS is 22$\times$ faster 
than NSIL and 306$\times$ faster than \MetaABD.
NeuralFastLAS performs each of its symbolic stages 
exactly once by utlising the opt-sufficient subset 
for network training. On the other hand, NSIL must 
perform the FastLAS solving stage once per epoch.
Similarly, \MetaABD\ must recompute the abduced labels 
for every epoch. 
Furthermore, \MetaABD\ needs many epochs to learn in the $a+b+c$ task,
as it struggles to converge, hence
the significantly higher training time. 
The results show that NeuralFastLAS scales to larger search spaces without extreme penalties to training time.

\section{Conclusion}

This paper presents NeuralFastLAS, a neuro-symbolic approach that 
jointly trains both components. NeuralFastLAS computes an opt-sufficient
subset of rules to train a neural network and then solves for the final 
hypothesis using the network predictions. The experiments show that NeuralFastLAS
achieves state-of-the-art accuracy in various tasks 
while being orders of magnitude faster than other neuro-symbolic methods.
It is restricted to solving tasks with non-recursive solutions, a limitation
imposed by FastLAS. Future work will be aimed at lifting this restriction
to expand the scope of tasks that can be solved.

\bibliographystyle{named}
\bibliography{ijcai23}

\begin{thebibliography}{}

\bibitem[\protect\citeauthoryear{Cropper and Muggleton}{2016}]{metagol}
Andrew Cropper and Stephen~H. Muggleton.
\newblock Metagol system.
\newblock https://github.com/metagol/metagol, 2016.

\bibitem[\protect\citeauthoryear{Cunnington \bgroup \em et al.\egroup
  }{2022}]{nsil}
Daniel Cunnington, Mark Law, Jorge Lobo, and Alessandra Russo.
\newblock Inductive learning of complex knowledge from raw data.
\newblock {\em arXiv}, abs/2205.12735, 2022.

\bibitem[\protect\citeauthoryear{Dai and Muggleton}{2021}]{wangzhou_ijcai}
Wang-Zhou Dai and Stephen Muggleton.
\newblock Abductive knowledge induction from raw data.
\newblock In Zhi-Hua Zhou, editor, {\em Proceedings of the Thirtieth
  International Joint Conference on Artificial Intelligence, {IJCAI-21}}, pages
  1845--1851. International Joint Conferences on Artificial Intelligence
  Organization, 8 2021.
\newblock Main Track.

\bibitem[\protect\citeauthoryear{Evans}{2022}]{aperception}
Richard Evans.
\newblock The apperception engine.
\newblock In Hyeongjoo Kim and Dieter Sch\"{o}necker, editors, {\em Kant and
  Artificial Intelligence}, pages 39--104. De Gruyter, 2022.

\bibitem[\protect\citeauthoryear{Hocquette and Muggleton}{2018}]{prior}
C{\'e}line Hocquette and Stephen Muggleton.
\newblock How much can experimental cost be reduced in active learning of
  agent strategies?
\newblock In Fabrizio Riguzzi, Elena Bellodi, and Riccardo Zese, editors, {\em
  Inductive Logic Programming}, pages 38--53, Cham, 2018. Springer
  International Publishing.

\bibitem[\protect\citeauthoryear{Koh \bgroup \em et al.\egroup }{2020}]{cbm}
Pang~Wei Koh, Thao Nguyen, Yew~Siang Tang, Stephen Mussmann, Emma Pierson, Been
  Kim, and Percy Liang.
\newblock Concept bottleneck models.
\newblock In Hal~Daumé III and Aarti Singh, editors, {\em Proceedings of the
  37th International Conference on Machine Learning}, volume 119 of {\em
  Proceedings of Machine Learning Research}, pages 5338--5348. PMLR, 13--18 Jul
  2020.

\bibitem[\protect\citeauthoryear{Law \bgroup \em et al.\egroup }{2015}]{ILASP}
Mark Law, Alessandra Russo, and Krysia Broda.
\newblock The {ILASP} system for learning answer set programs.
\newblock \url{www.ilasp.com}, 2015.

\bibitem[\protect\citeauthoryear{Law \bgroup \em et al.\egroup
  }{2020}]{fastlas}
Mark Law, Alessandra Russo, Elisa Bertino, Krysia Broda, and Jorge Lobo.
\newblock Fastlas: Scalable inductive logic programming incorporating
  domain-specific optimisation criteria.
\newblock {\em Proceedings of the AAAI Conference on Artificial Intelligence},
  34(03):2877--2885, Apr. 2020.

\bibitem[\protect\citeauthoryear{Law \bgroup \em et al.\egroup
  }{2021}]{fastlas2}
Mark Law, Alessandra Russo, Krysia Broda, and Elisa Bertino.
\newblock Scalable non-observational predicate learning in asp.
\newblock In Zhi-Hua Zhou, editor, {\em Proceedings of the Thirtieth
  International Joint Conference on Artificial Intelligence, {IJCAI-21}}, pages
  1936--1943. International Joint Conferences on Artificial Intelligence
  Organization, 8 2021.
\newblock Main Track.

\bibitem[\protect\citeauthoryear{Manhaeve \bgroup \em et al.\egroup
  }{2018}]{deepproblog}
Robin Manhaeve, Sebastijan Dumancic, Angelika Kimmig, Thomas Demeester, and Luc
  De~Raedt.
\newblock Deepproblog: Neural probabilistic logic programming.
\newblock In S.~Bengio, H.~Wallach, H.~Larochelle, K.~Grauman, N.~Cesa-Bianchi,
  and R.~Garnett, editors, {\em Advances in Neural Information Processing
  Systems}, volume~31. Curran Associates, Inc., 2018.

\bibitem[\protect\citeauthoryear{Muggleton \bgroup \em et al.\egroup
  }{2015}]{MIL}
Stephen~H. Muggleton, Dianhuan Lin, and Alireza Tamaddoni-Nezhad.
\newblock Meta-interpretive learning of higher-order dyadic datalog: predicate
  invention revisited.
\newblock {\em Machine Learning}, 100(1):49--73, Jul 2015.

\bibitem[\protect\citeauthoryear{Pascanu \bgroup \em et al.\egroup
  }{2012}]{rnn_difficult}
Razvan Pascanu, Tom{\'{a}}s Mikolov, and Yoshua Bengio.
\newblock Understanding the exploding gradient problem.
\newblock {\em CoRR}, abs/1211.5063, 2012.

\bibitem[\protect\citeauthoryear{Trask \bgroup \em et al.\egroup }{2018}]{nalu}
Andrew Trask, Felix Hill, Scott~E Reed, Jack Rae, Chris Dyer, and Phil Blunsom.
\newblock Neural arithmetic logic units.
\newblock In S.~Bengio, H.~Wallach, H.~Larochelle, K.~Grauman, N.~Cesa-Bianchi,
  and R.~Garnett, editors, {\em Advances in Neural Information Processing
  Systems}, volume~31. Curran Associates, Inc., 2018.

\bibitem[\protect\citeauthoryear{Tsamoura \bgroup \em et al.\egroup
  }{2021}]{efy}
Efthymia Tsamoura, Timothy Hospedales, and Loizos Michael.
\newblock Neural-symbolic integration: A compositional perspective.
\newblock {\em Proceedings of the AAAI Conference on Artificial Intelligence},
  35(6):5051--5060, 2021.

\bibitem[\protect\citeauthoryear{Xu \bgroup \em et al.\egroup
  }{2018}]{semantic_loss}
Jingyi Xu, Zilu Zhang, Tal Friedman, Yitao Liang, and Guy Van~den Broeck.
\newblock A semantic loss function for deep learning with symbolic knowledge.
\newblock In Jennifer Dy and Andreas Krause, editors, {\em Proceedings of the
  35th International Conference on Machine Learning}, volume~80 of {\em
  Proceedings of Machine Learning Research}, pages 5502--5511. PMLR, 10--15 Jul
  2018.

\bibitem[\protect\citeauthoryear{Yang \bgroup \em et al.\egroup
  }{2020}]{neurasp}
Zhun Yang, Adam Ishay, and Joohyung Lee.
\newblock Neurasp: Embracing neural networks into answer set programming.
\newblock In Christian Bessiere, editor, {\em Proceedings of the Twenty-Ninth
  International Joint Conference on Artificial Intelligence, {IJCAI-20}}, pages
  1755--1762. International Joint Conferences on Artificial Intelligence
  Organization, 7 2020.
\newblock Main track.

\end{thebibliography}

\end{document}